\pdfoutput=1

\documentclass[11pt]{article}

\usepackage{emnlp2021}

\usepackage{times}
\usepackage{latexsym}

\usepackage[T1]{fontenc}

\usepackage[utf8]{inputenc}

\usepackage{microtype}

\usepackage{amsmath}
\usepackage{mathtools}
\usepackage[ruled,vlined,noend]{algorithm2e}
\DeclareMathOperator*{\argmin}{argmin} 
\usepackage{graphicx}
\usepackage{mathtools}
\usepackage{bbm}
\usepackage{amsfonts}
\usepackage[inline]{enumitem}
\usepackage{tabularx}
\usepackage{arydshln}
\usepackage{multirow}
\usepackage{makecell}
\usepackage{xcolor}
\usepackage{soul}

\SetKwIF{If}{ElseIf}{Else}{if}{}{else if}{else}{end if}%
\SetKwFor{While}{while}{}{end while}%
\SetKwRepeat{Do}{do}{while}
\SetKw{KwGoTo}{go to}
\usepackage{mathtools}
\DeclarePairedDelimiter{\ceil}{\lceil}{\rceil}

\newif\ifcomments
\ifcomments
    \providecommand{\mi}[1]{{\protect\color{olive}{[MI: #1]}}}
    
    \providecommand{\jb}[1]{{\protect\color{blue}{[JB: #1]}}}
    \providecommand{\tw}[1]{{\protect\color{magenta}{[TW: #1]}}}
    \providecommand{\io}[1]{{\protect\color{pink}{[IO: #1]}}}
    \providecommand{\mh}[1]{{\protect\color{pink}{[MH: #1]}}}
    \providecommand{\ol}[1]{{\protect\color{orange}{[OL: #1]}}}
    \providecommand{\us}[1]{{\protect\color{brown}{[US: #1]}}}
    \providecommand{\yk}[1]{{\protect\color{magenta}{[yk: #1]}}}
\else
    \providecommand{\mi}[1]{}
    
    \providecommand{\jb}[1]{}
    \providecommand{\tw}[1]{}
    \providecommand{\io}[1]{}
    \providecommand{\mh}[1]{}
    \providecommand{\ol}[1]{}
    \providecommand{\us}[1]{}
    \providecommand{\yk}[1]{}
\fi

\newcommand\sE{\mathcal{E}}
\newcommand\sN{\mathcal{N}}
\newcommand\sS{\mathcal{S}}

\newcommand\sU{\mathcal{U}}
\newcommand\sW{\mathcal{W}}
\newcommand\sX{\mathcal{X}}
\newcommand\sY{\mathcal{Y}}
\newcommand\baseline{\textsc{Baseline}}
\newcommand\bertb{\textsc{BERT-Base\ }}
\newcommand\robertal{\textsc{RoBERTa-Large\ }}
\newcommand\nlang[1]{\emph{``#1''}}

\newcommand\randoff{\textsc{RandOff}}
\newcommand\randon{\textsc{RandOn}}
\newcommand\randonadv{\textsc{RandOn}}
\newcommand\randoffadv{\textsc{Rnd-OA}}
\newcommand\advoff{\textsc{AdvOff}}
\newcommand\advon{\textsc{AdvOn}}
\newcommand\bffoff{\textsc{BFFOff}}
\newcommand\bffon{\textsc{BFFOn}}
\newcommand\bff{\textsc{BFF}}
\newcommand\txfoff{\textsc{TxFOff}}
\newcommand\txfon{\textsc{TxFOn}}
\newcommand\txf{\textsc{TxF}}

\newcommand\freelb{\textsc{FreeLB}}

\newcommand\syn[1]{\textit{Syn($#1$)}}
\newcommand\commentout[1]{}
\newcommand{\NA}{-}
\newcommand\pert[1]{\textcolor{red}{\textbf{#1}}}

\title{Achieving Model Robustness through Discrete Adversarial Training}

\author{Maor Ivgi \\
  Tel-Aviv University \\
  \texttt{maorivgi@mail.tau.ac.il} \\\And
  Jonathan Berant \\
  Tel-Aviv University \\
  The Allen Institute for AI \\
  \texttt{joberant@cs.tau.ac.il} \\}

\begin{document}
\maketitle

\begin{abstract}
Discrete adversarial attacks are symbolic perturbations to a language input that preserve the output label but lead to a prediction error. While such attacks have been extensively explored for the purpose of \emph{evaluating} model robustness, their utility for \emph{improving robustness} has been limited to offline augmentation only. Concretely, given a trained model, attacks are used to generate perturbed (adversarial) examples, and the model is re-trained exactly once. In this work, we address this gap and leverage discrete attacks for \emph{online augmentation}, where adversarial examples are generated at every training step, adapting to the changing nature of the model. We propose (i) a new discrete attack, based on best-first search, and (ii) random sampling attacks that unlike prior work are not based on expensive search-based procedures. 
Surprisingly, we find that random sampling leads to impressive gains in robustness, outperforming the commonly-used offline augmentation, while leading to a speedup at training time of $\sim$10x. 
Furthermore, online augmentation with search-based attacks justifies the higher training cost, significantly improving robustness on three datasets. Last, we show that our new attack substantially improves robustness compared to prior methods.

\end{abstract}

\section{Introduction}

\begin{figure}[t]
\begin{center}
\centerline{\includegraphics[trim={0 0 0.8cm 0},clip,width=\linewidth]{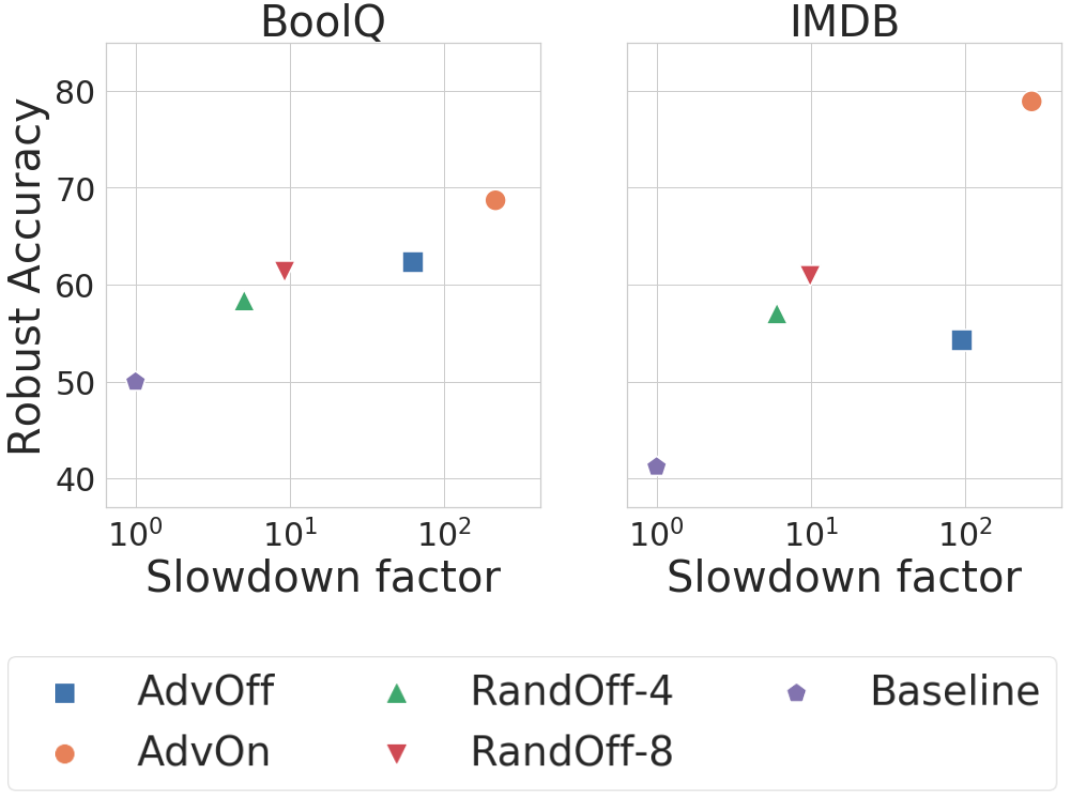}}
\caption[Robust accuracy vs. slowdown in training time]{Robust accuracy vs. slowdown in training time, comparing different methods to Baseline (purple pentagon); x-axis in logarithmic scale. The popular \advoff{} (blue squares, offline augmentation with adversarial example) is 10x slower than our simple augmentation of 4 (8) random samples (triangles, \randoff{}-4, \randoff{}-8) and achieves similar or worse robust accuracy. Our online augmentation of adversarial examples (\advon{}, yellow circles) significantly improves robust accuracy, but is expensive to train.
} 
\label{fig:intro}
\end{center}
\end{figure}

Adversarial examples are inputs that are slightly, but intentionally, perturbed to create a new example that is misclassified by a model \cite{szegedy2013intriguing}. 
Adversarial examples have attracted immense attention in machine learning \cite{goodfellow2014explaining,Carlini2017TowardsET,papernot2017practical} for two important, but separate, reasons.
First, they are useful for \emph{evaluating} model robustness, and have revealed that current models are over-sensitive to minor perturbations.
Second, adversarial examples can \emph{improve} robustness: training on adversarial examples reduces the brittleness and over-sensitivity of deep learning models to such perturbations \cite{alzantot2018generating,jin2020bert,li2020bert,lei2018discrete,wallace2019universal,zhang2020attacks, garg2020bae, si2020benchmarking, goel2021robustness}.

Training and evaluating models with adversarial examples has had considerable success in computer vision, with gradient-based techniques like FGSM \cite{goodfellow2014explaining} and PGD \cite{madry2017towards}. In computer vision, adversarial examples can be constructed by considering a continuous space of imperceptible perturbations around image pixels. Conversely, language is discrete, and any perturbation is perceptible. Thus, robust models must be invariant to input modifications that preserve semantics, such as synonym substitutions \cite{alzantot2018generating,jin2020bert}, paraphrasing \cite{tan2020s}, or typos \cite{huang2019achieving}.

Due to this property of language, ample work has been dedicated to developing discrete attacks that generate adversarial examples through combinatorial optimization  \cite{alzantot2018generating,ren2019generating,jin2020bert,zhou2020defense,zang2020word} . For example, in sentiment analysis, it is common to consider the space of all \emph{synonym substitutions}, where an adversarial example for an input \nlang{Such an amazing movie!} might be \nlang{Such an extraordinary film} (Fig.~\ref{fig:setup}).
This body of work has mostly focused on \emph{evaluating} robustness, rather than \emph{improving it}, which naturally led to the development of complex combinatorial search algorithms, whose goal is to find adversarial examples in the exponential space of perturbations. 

In this work, we address a major research gap in current literature around \emph{improving} robustness with discrete attacks.  Specifically, past work \cite{alzantot2018generating,ren2019generating,jin2020bert} only considered \emph{offline augmentation}, where a discrete attack is used to generate adversarial examples and the model is re-trained exactly once with those examples. This ignores \emph{online augmentation}, which had success in computer vision \cite{kurakin2016adversarial,perez2017effectiveness,madry2017towards}, where adversarial examples are generated in each training step, adapting to the changing model. Moreover, simple data augmentation techniques, such as randomly sampling from the space of synonym substitutions and adding the generated samples to the training data have not been investigated and compared to offline adversarial augmentation. We address this lacuna and systematically compare online augmentation to offline augmentation, as well as to simple random sampling techniques. To our knowledge, we are the first to evaluate \emph{online augmentation} with discrete attacks on a wide range of NLP tasks. 
Our results show that online augmentation leads to significant improvement in robustness compared to prior work and that simple random augmentation achieves comparable results to the common offline augmentation at a fraction of the complexity and training time.

Moreover, we present a new search algorithm for finding adversarial examples, \emph{Best-First search over a Factorized graph (BFF)},
which alleviates the greedy nature of previously-proposed algorithms. BFF improves search by incorporating backtracking, and allowing to re-visit previously-discarded search paths, once the current one is revealed to be sub-optimal.

We evaluate model robustness on three datasets: BoolQ \cite{clark2019boolq}, IMDB \cite{maas-EtAl:2011:ACL-HLT2011}, and SST-2 \cite{socher2013recursive}, which vary in terms of the target task (question answering and sentiment analysis) and input length. 
Surprisingly, we find across different tasks (Fig.~\ref{fig:intro}) that augmenting each training example with 4-8 random samples from the synonym substitution space performs as well as (or better than) the commonly used offline augmentation, while being simpler and 10x faster to train. Conversely, online augmentation makes better use of the extra computational cost, and substantially improves robust accuracy compared to offline augmentation. Additionally, our proposed discrete attack algorithm, BFF, outperforms prior work by a wide margin. 
Our data and code are available at \url{https://github.com/Mivg/robust_transformers}.

\section{Problem Setup and Background}
\label{sec:setup}

\begin{figure}[t]
\begin{center}
\centerline{
\includegraphics[width=\linewidth]{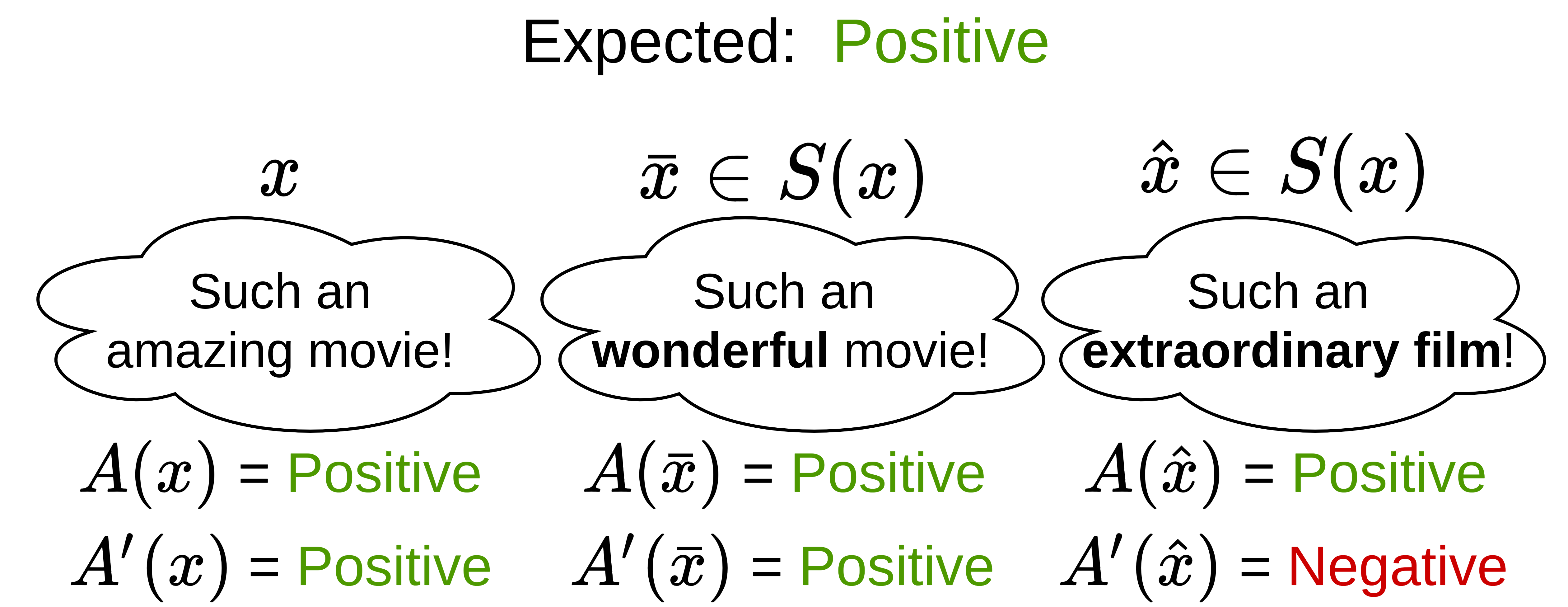}
}
\caption[Example of model non-robustness to perturbations]{Given a movie review $x$, the model $A$ is robust to a set of perturbations, while $A'$ is not.}
\label{fig:setup}
\end{center}
\end{figure}

\paragraph{Problem setup}  We focus in this work on the  supervised classification setup, where given a training set $\{x_j, y_j\}_{j=1}^N$ sampled from $\mathcal{X} \times \mathcal{Y}$, our goal is to learn a mapping $A:\mathcal{X}\rightarrow \mathcal{Y}$ that achieves high accuracy on held-out data sampled from the same distribution. Moreover, we want the model $A$ to be \emph{robust}, i.e., invariant to a set of  pre-defined label-preserving perturbations to $x$, such as synonym substitutions. Formally,
for any natural language input $x$, a discrete \emph{attack space} of label-preserving perturbations $\sS(x) \subset \sX$ is defined. Given a labeled example $(x,y)$, a model $A$ is robust w.r.t $x$, if $A(x) = y$ and for any $\bar{x} \in \sS(x)$, the output $A(\bar{x}) = A(x)$. An example $\bar{x} \in \sS(x)$ such that $A(\bar{x}) \neq A(x)$ is called an \emph{adversarial example}. We assume $A$ provides not only a prediction but a distribution $p_A(x) \in \Delta^{|\sY|}$ over the possible classes, where $\Delta$ is the simplex, and denote the probability $A$ assigns to the gold label by $[p_A(x)]_y$.
Fig.~\ref{fig:setup} shows an example from sentiment analysis, where a model $A$ is robust, while $A'$ is not w.r.t $x$.

Robustness is evaluated with \emph{robust accuracy}  \cite{Tsipras2019RobustnessMB}, i.e., the fraction of examples a model is robust to over some held-out data. Typically, the size of the attack space $\sS(x)$ is exponential in the size of $x$ and it is not feasible to enumerate all perturbations. Instead, an upper bound is estimated by searching for a set of adversarial attacks, i.e., ``hard'' examples in $\sS(x)$ for every $x$, and estimating robust accuracy w.r.t to that set. 

\paragraph{Improving robustness with discrete attacks} Since language is discrete, a typical approach for \emph{evaluating} robustness is to use combinatorial optimization methods to search for adversarial examples in the attack space $\sS(x)$.
This has been repeatedly shown to be an effective attack method on pre-trained models \cite{alzantot2018generating,lei2018discrete, ren2019generating,li2020bert,jin2020bert,zang2020word}.
However, in terms of \emph{improving} robustness, discrete attacks have thus far
been mostly used with offline augmentation (defined below) and have led to limited robustness gains. In this work, we examine the more costly but potentially more beneficial online augmentation.

\paragraph{Offline vs. online augmentation}
Data augmentation is a common approach for improving generalization and robustness, where variants of training examples are automatically generated and added to the training data \cite{simard1998transformation}. Here, discrete attacks can be used to generate these examples. We consider both \emph{offline} and \emph{online} data augmentation and focus on improving robustness with adversarial examples.

Given a training set $\{(x_j, y_j)\}_{j=1}^N$, \emph{offline data augmentation} involves (a) training a model $A$ over the training data, (b) for each training example $(x_j, y_j)$, generating a perturbation w.r.t to $A$ (using some discrete attack) and labeling it with $y_j$, and (c) training a new model over the union of the original training set and the generated examples. This is termed \emph{offline} augmentation because examples are generated with respect to a fixed model $A$.

\emph{Online data augmentation} is this setup, examples are generated at training time w.r.t the current model $A$.
This is more computationally expensive, as examples must be generated during training and not as pre-processing, but examples can adapt to the model over time.
In each step, half the batch contains examples from the training set, and half are adversarial examples generated by some discrete attack w.r.t to the model's current state.

Online augmentation has been used to improve robustness in NLP with gradient-based approaches \cite{jia2019certified,shi2020robustness,zhou2020defense}, but to the best of our knowledge has been overlooked in the context of discrete attacks. In this work, we are the first to propose model-agnostic online augmentation training, which uses automatically generated \emph{discrete adversarial attacks} to boost overall robustness in NLP models. 

\section{The Attack Space}
\label{sec:attack_space}

An attack space for an input with respect to a classification task can be intuitively defined as the set of label-preserving perturbations over the input.
A popular attack space $\sS(x)$, which we adopt, is the space of \emph{synonym substitutions}
\cite{alzantot2018generating, ren2019generating}.
Given a synonym dictionary that provides a set of synonyms \syn{w} for any word $w$, the attack space $\sS_\textit{syn}(x)$ for an utterance $x = (w_1, \dots, w_n)$ contains all utterances that can be obtained by replacing a word $w_i$ (and possibly multiple words) with one of their synonyms. Typically, the number of words from $x$ allowed to be substituted is limited to be no more than $D = \ceil{d\cdot|x|}$, where $d \in \{0.1, 0.2\}$ is a common choice.

Synonym substitutions are context-sensitive, i.e., substitutions might only be appropriate in certain contexts. For example, in Fig.~\ref{fig:search-alg}, replacing the word \nlang{like} with its synonym \nlang{similar} (red box) is invalid, since \nlang{like} is a verb in this context. Consequently, past work \cite{ren2019generating,jin2020bert} filtered $\sS_{\textit{syn}}(x)$ using a context-sensitive filtering function $\Phi_x(w_i,\bar{w}_i) \in \{0,1\}$, which determines whether substituting a word $w_i$ from the original utterance $x$ with its synonym $\bar{w}_i$ is valid in a particular context.
For instance, an external model can check whether the substitution maintains the part-of-speech, and whether the overall semantics is maintained.
We define the \emph{filtered synonyms substitutions space} $\sS_\Phi(x)$ as the set that includes all utterances $\bar{x}$ that can be generated through a sequence of no more than $D$ 
single-word substitutions from the original utterance that are valid according to $\Phi(\cdot,\cdot)$. 
In \S\ref{subsec:attack_space}, we describe the details of the synonym dictionary and function $\Phi$.

\section{Best-first Search Over a Factorized Graph}
\label{sec:attacks}

\begin{figure}[t]
\begin{center}
\centerline{\includegraphics[width=\linewidth]{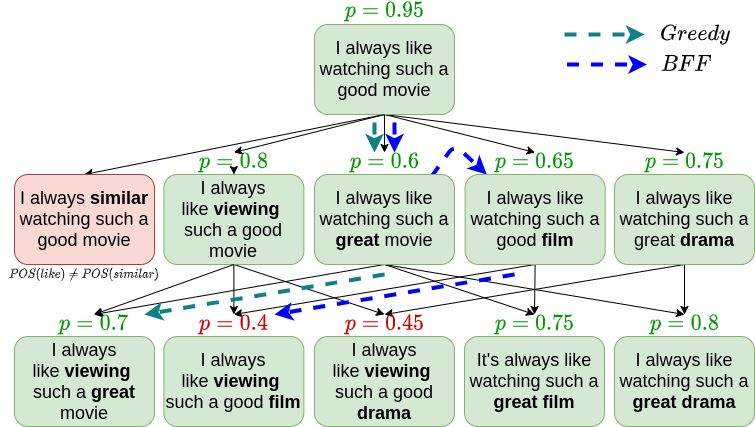}}
\caption[Example of an attack space and attack strategies over it]{Example of an attack space, and the paths taken by a greedy algorithm and best-first search. An adversarial example has a probability $p < 0.5$ for the gold positive label.} 
\label{fig:search-alg}
\end{center}
\end{figure}

Searching over the attack space $\sS_\Phi(x)$  can be naturally viewed as a search problem over a  directed acyclic graph (DAG), $G = (\sU,\sE)$, where each node $u_{\bar{x}} \in \sU$ is labeled by an utterance $\bar{x}$, and edges $\mathcal{E}$ correspond to single-word substitutions, valid according to $\Phi(\cdot)$. The graph is directed and acyclic, since only substitutions of words from the original utterance $x$ are allowed (see Fig.~\ref{fig:search-alg}).
Because there is a one-to-one mapping from the node $u_{\bar{x}}$ to the utterance $\bar{x}$, we will use the latter to denote both the node and the utterance.

Discrete attacks use search algorithms to find an adversarial example in $\sS(x)$. 
The search is guided by a heuristic scoring function $s_A(x) \coloneqq [p_A(x)]_y$, where the underlying assumption is that utterances that give lower probability to the gold label are closer to an adversarial example.
A popular choice for a search algorithm in NLP is greedy search, illustrated in Fig.~\ref{fig:search-alg}. Specifically, one holds in step $t$ the current node $x_t$, where $t$ words have been substituted in the source node $x_0 = x$. Then, the model $A(\cdot)$ is run on the \emph{frontier}, that is, all out-neighbor nodes $\sN(x_t) = \{\hat{x}_{t+1} \mid (x_t, \hat{x}_{t+1}) \in \sE\}$, and the one that minimizes the heuristic scoring function is selected:
$x_{t+1} \coloneqq \argmin_{\hat{x} \in \sN(x_t)} s_A(\hat{x})$. 

While greedy search has been used for character-flipping \cite{ebrahimi2017hotflip}, it is ill-suited in the space of synonym substitutions. The degree of nodes is high -- assuming $n_\text{rep}$ words can be replaced in the text, each with $K$ possible synonyms, then the out degree is $O(n_\text{rep} \cdot K)$. This results in an infeasible number of forward passes through the attacked model even for a small number of search iterations.

To enable effective search through the search space, we (a) factorize the graph such that the out-degree of nodes is lower, and (b) use a best-first search algorithm. We describe those next.

\paragraph{Graph factorization}
To reduce the out-degree of a node in the search space and thus improve its efficiency, we can split each step into two. First, choose a position to substitute in the utterance; Second, choose a substitution for that position. This reduces the number of evaluations of $A$ per step from $O(n_\text{rep} \cdot K)$ to $O(n_\text{rep} + K)$. To estimate the score of a position $i$, one can mask the word $w_i$ with a mask token $\tau$ and measure $s_A(x_{w_i \rightarrow \tau})$, where $x_{w_i \rightarrow \tau}$ is the utterance $x$ where the word in position $i$ is replaced by the mask $\tau$.

We can describe this approach as search over a bi-partite DAG $\hat{G}=(\sU\cup \mathcal{W},\hat{\sE})$. The nodes $\sU$ are utterances like in $G$, and the new nodes are utterances with a single mask token $\sW = \{ \bar{x}_{w_i \rightarrow \tau} \mid \bar{x} \in \sS(x) \land w_i \text{ is a word in } x \}$. The edges comprise two types: $\hat{\sE} = \sE_1 \cup \sE_2$. The edges $\sE_1$ are from utterances to masked utterances: $\sE_1 = \{(\bar{x}, \bar{x}_{w_i \rightarrow \tau})\}  \subset \sU \times \sW$, and $\mathcal{E}_2 = \{ (\bar{x}_{w_i \rightarrow \tau}, \bar{x}_{w_i \rightarrow w_\mathit{syn}}) \}  \subset \sW \times \sU$, where $w_\mathit{syn}\in \syn{w_i}$. In Figure~\ref{fig:search-alg}, the two rightmost nodes in each row would be factorized together as they substitute the same word, and the algorithm will evaluate only one of them to estimate the potential benefit of substituting \nlang{movie}.

\paragraph{Best-first search}
\label{subsec:bff}

A factorized graph makes search possible by reducing the out-degree of nodes. However, greedy search is still sub-optimal. This is since it relies on the heuristic search function to be a good estimate of the distance to an adversarial example, which can often be false. Consider the example in Fig.~\ref{fig:search-alg}. The two adversarial examples (with $p=0.4$ or $p=0.45$) are not reachable from the best node after the first step ($p=0.6$), only from the second-best  ($p=0.65$).

Best-first search \cite{10.5555/525} overcomes this at a negligible cost, by holding a min-heap over the nodes of the frontier of the search space (Alg.~\ref{alg:backtracking}). In each step, we pop the next utterance, which assigns the lowest probability to the gold label, and push all neighbors into the heap. When a promising branch turns out to be sub-optimal, search can resume from an earlier node to find a better solution, as shown in the blue path in Figure~\ref{fig:search-alg}. To bound the cost of finding a single adversarial example, we bound the number of forward passes through the model $A$ with a budget parameter $B$.
To further reduce ``greedyness'', search can use a beam by popping more than one node in each step, expanding all their neighbors and pushing the result back to the heap.
Our final approach uses \textbf{B}est-\textbf{F}irst search over a \textbf{F}actorized graph, and is termed \textbf{BFF}.

\begin{algorithm}
\footnotesize
\SetAlgoLined
    \SetKwInOut{Input}{input}
    \SetKwInOut{Output}{output}
    \Input{model A, factorized graph $G$, utterance $x$.}
    $\mathtt{heap} \leftarrow \{(x, s_A(X)\}$ \\
    $x^* \leftarrow x$ \\ 
    \While{$\vert \mathtt{heap} \vert > 0$ and budget $B$ not exhausted:}
    {
        $\bar{x} \leftarrow \mathtt{heap.pop()}$ \\
        $x^* \leftarrow \argmin_{\hat{x}\in\{\bar{x}, x^*\}}{A(\hat{x})}$ \\
        \lIf{$A(x^*)\ne y$}{\textbf{break}}
        \For{$\hat{x} \in \mathcal{N}(\bar{x})$} {
            $\mathtt{heap.push}(\hat{x},s_A(\hat{x}))$
        }
    }
    \Return $x^*$ 
 \caption{BFF 
 }
 \label{alg:backtracking}
\end{algorithm}
\section{Experiments}
\label{sec:experiments}

We conduct a thorough empirical evaluation of model robustness across a wide range of attacks and training procedures.

\subsection{Experimental Setup}

To evaluate our approach over diverse settings, we consider three different tasks: \emph{text classification}, \emph{sentiment analysis} and \emph{question answering}, two of which contain long passages that result in a large attack space (see Table~\ref{table:attack-space}). 
\begin{enumerate}[leftmargin=*,itemsep=0pt,topsep=0pt]
    \item \textbf{SST-2}: Based on the the Stanford sentiment treebank~\cite{socher2013recursive}, SST-2 is a binary (positive/negative) classification task containing 11,855 sentences describing movie reviews. SST-2 has been frequently used for evaluating robustness.
    \item \textbf{IMDB} \cite{maas-EtAl:2011:ACL-HLT2011}:
    A binary (positive/negative) text classification task, containing 
    50K reviews from IMDB. Here, passages are long and thus the attack space is large (Table~\ref{table:attack-space}). 
    \item \textbf{BoolQ} \cite{clark2019boolq}: contains 16,000 yes/no questions over Wikipedia paragraphs. This task is perhaps the most interesting, because the attack space is large and answering requires global passage understanding. We allow word substitutions in the paragraph only and do not substitute nouns, verbs, or adjectives that appear in the question to avoid non-label-preserving perturbations. Further details can be found in App.~\ref{appendix:attack-space}.
\end{enumerate}

\noindent
\textbf{Models}
We consider a wide array of models and evaluate both their downstream accuracy and robustness. In all models, we define a budget of $B=1000$, which specifies the maximal number of allowed forward passes through the model for finding an adversarial example. All results are an average of 3 runs. 

To demonstrate the effectiveness of \bff{} for both robustness evaluation as well as adversarial training, we compare it to a recent state-of-the-art discrete attack, \textsc{TextFooler} \cite{jin2020bert}, which we denote in model names below by the prefix \textsc{TxF}. The models compared are:

\begin{itemize}[leftmargin=0pt,itemsep=0pt,topsep=0pt]
    \item \baseline: we fine-tune a pretrained language model on the training set. We use \bertb \cite{devlin2018bert} for IMDB/SST-2 and \robertal \cite{liu2019roberta} for BoolQ. These baselines are on par with current state-of-the-art to demonstrate the efficacy of our method.
    \item \bffoff/\txfoff{} Offline augmentation with the \bff{} or \textsc{TextFooler} attacks.
    \item \bffon/\txfon{} Online augmentation with the \bff{} or \textsc{TextFooler} attacks.
    \item \randoff-$L$: We compare search-based algorithms to a simple and efficient approach that does not require any forward passes through the model $A$. Specifically, we randomly sample $L$ utterances from the attack space for each example (without executing $A$) and add them to the training data.
    \item \randon{}: A random sampling approach that does use the model $A$. Here, we sample $B$ random utterances, pass them through $A$, and return the attack that resulted in lowest model probability.
    \item \freelb{}: For completeness, we also consider \freelb{} \cite{zhu2019freelb}, a popular gradient-based approach for improving robustness, which employs \emph{virtual adversarial training} (see \S\ref{sec:related}). This approach uses online augmentation, where examples are created by taking gradient steps w.r.t the input embeddings to maximize the model's loss.
    Other gradient-based approaches (e.g., certified robustness) are not suitable when using pre-trained transformers, which we further discuss in \S\ref{sec:related}.
  
\end{itemize}

In a parallel line of work, \newcite{garg2020bae} and \newcite{li2020bert} used pre-trained language models to both \emph{define} an attack space and to \emph{generate} high-fidelty attacks in that space. while successful, these approaches are not suitable for our setting, due to the strong coupling between the attack strategy and the attack space itself. We further discuss this in \S\ref{subsec:bert-attack}

\noindent
\textbf{Evaluation}
\label{subsec:evalaution}
We evaluate models on their downstream accuracy, as well as on robust accuracy, i.e. the fraction of examples against which the model is robust. Since exact robust accuracy is intractable to compute due to the exponential size of the attack space, we compute an upper-bound by attacking each example with both \bff{} and \textsc{TextFooler} (\txf{}) with a budget of $B=2000$.
An example is robust if we cannot find an utterance where the prediction is different from the gold label. We evaluate robust accuracy on 1000/1000/872 samples from the development sets of BoolQ/IMDB/SST-2.

\subsection{Attack Space}
\label{subsec:attack_space}
Despite the myriad of works on discrete attacks, 
an attack space for synonym substitutions has not been standardized. While all past work employed a synonym dictionary combined with a $\Phi(\cdot,\cdot)$ filtering function (see \S\ref{sec:attack_space}), the particular filtering functions vary. 
When examining the attack space proposed in \txf{}, we observed that attacks result in examples that are difficult to understand or are not label-preserving. Table~\ref{table:bad-utterances} in App.~\ref{app:attack-prior-work} shows several examples. For instance, in sentiment classification, the attack replaced \nlang{compelling} with \nlang{unconvincing} in the sentence \nlang{it proves quite unconvincing as an intense , brooding character study} which alters the meaning and the sentiment of the sentence. Therefore, we use a more strict definition of the filtering function and conduct a user study to verify it is label-preserving.

Concretely, we use the synonym dictionary from \newcite{alzantot2018generating}. We determine if a word substitution is context-appropriate by computing all single-word substitutions ($n_\text{rep} \cdot K$) and disallowing those that change the POS tag according to spaCy \cite{spacy} or increase perplexity according to GPT-2 \cite{radford2019language} by more than 25\%. Similar to \newcite{jin2020bert}, we also filter out synonyms that are not semantics-preserving according to the \textsc{USE} \cite{cer2018universal} model. The attack space includes any combination of allowed single-word substitutions, where the fraction of allowed substitutions is $d=0.1$. Implementation details are in App.~\ref{appendix:attack-space}. We find that this ensemble of models reduces the number of substitutions that do not preserve semantics and are allowed by the filtering function.

\begin{table}[t]
\centering
\footnotesize
\begin{tabular}{l@{\extracolsep{6pt}~~}cccc} 
\Xhline{2\arrayrulewidth}
& { $\vert x\vert$} & $n_\text{rep}$ & { $\vert \syn{w}\vert$} & { $\vert S_\phi(x)\vert$} \\
\hline
SST-2           & $8.9$    & $2.7$  & $2.4$  & $27.7$   \\
IMDB            & $242.4$  & $97.3$ & $3.6$  & $2.27\times 10^{64}$  \\
BoolQ$^\dagger$ & $97.7$   & $38.7$ & $3.6$  & $3.64\times 10^{25}$  \\
\Xhline{2\arrayrulewidth}
\end{tabular}
\caption{Statistics on datasets and the size of attack space. We show the average number of words per utterance $|x|$, the average number of words with substitutions $n_\text{rep}$, average number of synonyms per replaceable word, and an estimation of the attack space size.}
\label{table:attack-space}
\end{table}

We check the validity of our more restrictive attack space with a user study, where we verify that our attack space is indeed label-preserving. The details of the user study are in \S\ref{subsec:userstudy}.

\subsection{Robustness Results}
\label{subsec:results}

\begin{table*}[t]
\centering
\footnotesize

\begin{tabular}{@{\extracolsep{6pt}}lccccccccc@{}} 

\Xhline{2\arrayrulewidth}
\multirow{2}{*}{\textbf{Model}} & \multicolumn{3}{c}{\textbf{Accuracy}} & \multicolumn{3}{c}{\textbf{Robust Accuracy}} & \multicolumn{3}{c}{\textbf{Slowdown}} \\\cline{2-4}\cline{5-7}\cline{8-10}

& {\footnotesize \textbf{SST-2}} & {\footnotesize \textbf{IMDB}} & {\footnotesize \textbf{BoolQ}} & {\footnotesize \textbf{SST-2}} & {\footnotesize \textbf{IMDB}} & {\footnotesize \textbf{BoolQ}} & {\footnotesize \textbf{SST-2}} & {\footnotesize \textbf{IMDB}} & {\footnotesize \textbf{BoolQ}} \\ \cline{2-2}\cline{3-3}\cline{4-4}\cline{5-5}\cline{6-6}\cline{7-7}\cline{8-8}\cline{9-9}\cline{10-10}

Baseline          & $91.9$ & $93.4$                & $84.5$
                  & $80.5$          & $41.2$                & $50.0$  
                  & $\times 1$      & $\times 1$            & $\times 1$ \\

\freelb{}           & $\textbf{92.5}$ & $93.9$                & $85.5$
                  & $82.1$          & $62.5$                & $55.8$  
                  & $\times 1.8$      & $\times 1.8$            & $\times 3.9$ \\
\hline\hline

\randoff-1        & $91.9$ & $93.5$                & $85.6$
                  & $83.5$          & $50.3$                & $52.2$
                  & $\times 1.9$    & $\times 1.5$          & $\times 2.1$ \\

\randoff-4        & $91.6$          & $93.7$                & $85.5$
                  & $83.6$          & $57.0$                & $58.4$ 
                  & $\times 3.8$    & $\times 4.5$          & $\times 5.1$ \\

\randoff-8        & $91.1$          & $93.8$                & $86.1$
                  & $83.3$          & $60.9$                & $61.3$ 
                  & $\times 5.4$    & $\times 8.0$          & $\times 9.3$ \\

\randoff-12       & $91.5$          & $93.7$                & $85.8$
                  & $84.2$          & $60.1$                & $63.0$ 
                  & $\times 6.3$    & $\times 11.5$         & $\times 13.2$ \\
\hline

\txfoff{}         & $91.2$          & $93.4$                & $\textbf{86.5}$ 
                  & $83.5$          & $49.0$                & $61.5$ 
                  & $\times 3.0$    & $\times 56.1$         & $\times 8.6$  \\ 

\bffoff{}         & $91.8$          & $93.7$                & $85.8$
                  & $84.6$          & $54.3$                & $62.3$ 
                  & $\times 5.4$    & $\times 60.0$         & $\times 63.2$\\

\hline\hline

\randon{}         & $91.7$          & $94.1$                & $85.6$
                  & $84.9$          & $68.5$                & $66.0$ 
                  & $\times 14.8$   & $\times 249.3$        & $\times 280.4$ \\ 

\txfon{}          & $91.3$          & $93.8$                & $86.0$ 
                  & $84.0$          & $67.4$                & $65.3$
                  & $\times 3.9$    & $\times 58.0$         & $\times 28.1$  \\ 

\bffon{}          & $91.7$          & $\textbf{94.2}$   & $\textbf{86.5}$
                  & $\textbf{85.3}$ & $\textbf{78.9}$   & $\textbf{68.7}$ 
                  & $\times 21.1$   & $\times 270.7$    & $\times 215.9$ \\
\Xhline{2\arrayrulewidth}
\end{tabular}
\caption{Accuracy on the evaluation set, robust accuracy, and slowdown in model training for all datasets.}
\label{table:results}
\end{table*}

Table~\ref{table:results} shows accuracy on the development set, robust accuracy, and slowdown compared to \baseline{} for all models and datasets. For downstream accuracy, training for robustness either maintains or slightly increases downstream accuracy. This is not the focus of this work, but is indeed a nice side-effect.
For robust accuracy, discrete attacks substantially improve robustness: $80.5 \rightarrow 85.3$ on SST-2, $41.2 \rightarrow 78.9$ on IMDB, and $50.0 \rightarrow 68.7$ on BoolQ, closing roughly half the gap from downstream accuracy.

Comparing different attacks, online augmentation (\bffon{}), which has been overlooked in the context of discrete attacks, leads to dramatic robustness gains compared to other methods, but is slow to train -- 20-270x slower than \baseline{}. This shows the importance of  continuous adaptation to the current vulnerabilities of the model.

Interestingly, adding offline random samples (\randoff$-L$) consistently improves robust accuracy, and using $L=12$ leads to impressive robustness gains without executing $A$ at all, outperforming \bffoff{} in robust accuracy, and being $\sim$5x faster on IMDB and BoolQ. 
Moreover, random sampling is trivial to implement, and independent from the attack strategy.
Hence, the common practice of using offline augmentation with search-based attacks, such as \bffoff{}, seems misguided, and a better solution is to use random sampling. Online random augmentation obtains impressive results, not far from \bffon{}, without applying any search procedure, but is very slow, since it uses the entire budget $B$ in every example.

Comparing \bff{} to \txf{}, we observe that \bff{}, which uses best-first search, outperforms \txf{}
in both the online and offline setting. 
Last \freelb{}, which is based on virtual adversarial training, improves robust accuracy at a low computational cost, but is dramatically outperformed by discrete search-based attacks, including \bff{}.

\begin{table}[t]
\centering
\resizebox{\columnwidth}{!}{
\begin{tabular}{@{\extracolsep{0.1pt}}lcccccccc@{}}
\Xhline{2\arrayrulewidth}

\multirow{2}{*}{\textbf{Model}}  & \multicolumn{4}{c}{\footnotesize \textbf{IMDB}} & \multicolumn{4}{c}{\footnotesize \textbf{BoolQ}} \\\cline{2-5}\cline{6-9}

& {\footnotesize \textbf{Rand}} & {\footnotesize \textbf{\txf{}}} & {\footnotesize \textbf{\bff{}}} & {\footnotesize \textbf{Gen}} & {\footnotesize \textbf{Rand}} & {\footnotesize \textbf{\txf{}}} & {\footnotesize \textbf{\bff{}}}  & {\footnotesize \textbf{Gen}} \\ \cline{2-2}\cline{3-3}\cline{4-4}\cline{5-5}\cline{6-6}\cline{7-7}\cline{8-8}\cline{9-9}

Baseline      & $73.1$          & $70.2$ & $49.9$ & $54.1$              & $62.1$ & $67.7$ & $50.2$  & $52.0$ \\ \hline\hline
\randoffadv{} & $74.8$          & $74.7$ & $52.9$ & $59.1$               & $70.9$ & $72.0$ & $59.4$ & $62.0$ \\ 
\txfoff{}     & $67.7$          & $77.5$ & $52.5$ & $56.7$               & $71.0$ & $75.0$ & $61.5$  & $63.4$ \\
\bffoff{}     & $75.4$          & $76.9$ & $58.6$ & $64.1$               & $70.9$ & $74.8$ & $64.7$ & $65.2$ \\
\hline\hline
\randonadv{}  & $\textbf{87.0}$ & $76.4$ & $68.5$ & $79.6$               & $71.5$ & $72.6$ & $60.1$ & $67.5$ \\ 
\txfon{}      & $81.1$          & $84.2$ & $69.7$ & $73.7$               & $73.4$ & $74.8$ & $65.3$ & $67.4$\\
\bffon{}      & $\textbf{87.0}$          & $\textbf{84.9}$ & $\textbf{79.0}$  & $\textbf{81.9}$     & $\textbf{75.1}$ & $\textbf{76.1}$ & $\textbf{69.0}$ & $\textbf{70.3}$  \\
\Xhline{2\arrayrulewidth}
\end{tabular}
}
\vskip 0.1in
\caption{Robust accuracy of different robust models w.r.t particular discrete attacks. \randoffadv{} is offline augmentation with a random attack and $B=1000$. \emph{Gen} is our implementation of the Genetic Attack by \newcite{alzantot2018generating}.
}
\label{table:results-strategies}
\end{table}

To summarize, random sampling leads to significant robustness gains at a small cost, outperforming the commonly used offline augmentation. Online augmentation leads to the best robustness, but is more expensive to train.

\subsection{Robustness across Attack Strategies}
A natural question is whether a model trained for robustness with an attack (e.g., \bff{}) is robust w.r.t to examples generated by other attacks, which are potentially uncorrelated with them. To answer that, we evaluate the robustness of our models to attacks generated by \bff{}, \txf{}, and random sampling. Moreover, we evaluate robustness to a genetic attack, which should not be correlated with \bff{} and \txf{}: we re-implement the genetic attack algorithm from  \newcite{alzantot2018generating} (details in \ref{appendix:gen-attack-implmentation-details}), and examine the robustness of our model to this attack. All attacks are with a budget of $B=2000$.

Table~\ref{table:results-strategies} shows the result of this evaluation.
We observe that \bffon{} obtains the highest robust accuracy results w.r.t to all attacks: \bff{}, \txf{}, random sampling, and a genetic attack. In offline augmentation, we observe again that \bffoff{} obtains good robust accuracy, higher or comparable to all other offline models for any attack strategy. This result highlights the generality of \bff{} for improving model robustness.

\subsection{Success Rate Results}
\label{subsec:search-alg-eval}
To compare the different attacks proposed in \S\ref{sec:attacks}, we analyze the \emph{success rate} against \baseline{}, i.e., the proportion of examples for which an attack finds an adversarial example as a function of the budget $B$.

\begin{figure}[t]
\begin{center}
\centerline{\includegraphics[width=\linewidth]{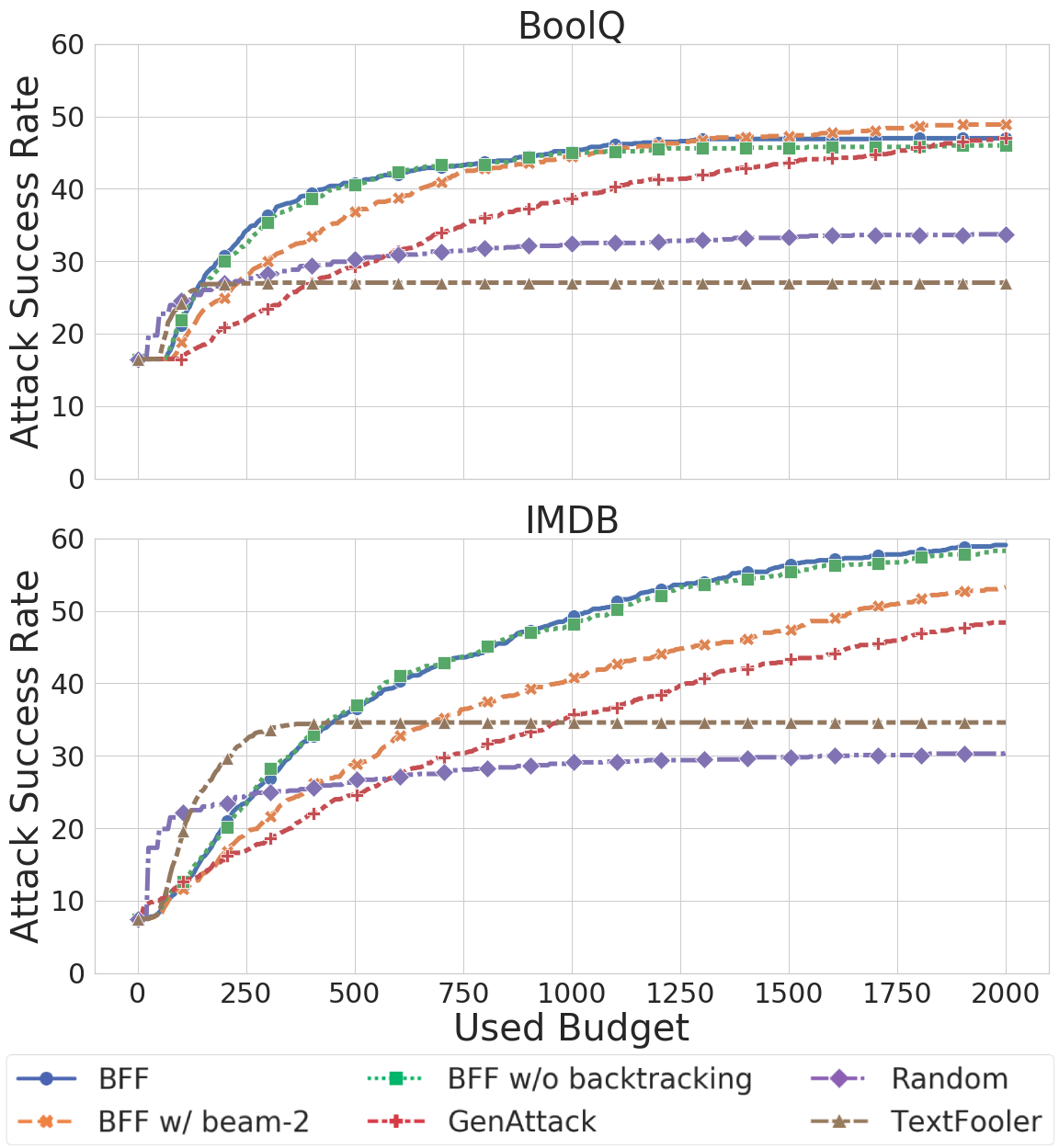}}
\caption{Success rate  of different attacks against BoolQ/IMDB \baseline{} as a function of the budget.}
\label{fig:attack-results}
\end{center}
\end{figure}

Fig.~\ref{fig:attack-results} compares the success rate of different attacks. We observe that BFF-based attacks have the highest success rate after a few hundred executions.
\textsc{TextFooler} performs well at first, finding adversarial examples for many examples, but then its success plateaus.
Similarly, a random approach, which ignores the graph structure,  starts with a relatively high success rate, as it explores far regions in the graph, but fails to properly utilize its budget and then falls behind.

\bff{} combines backtracking with graph factorization. When removing backtracking, i.e., greedy search over the factorized graph, success rate decreases, especially in BoolQ. Greedy search without graph factorization leads to a low success rate due to the large number of neighbors of each node, which quickly exhausts the budget. Moreover, looking at \bff{} with beam size 2 (popping 2 items from the heap in each step) leads to lower performance when the budget $B \leq 2000$, as executions are expended on less promising utterances, but could improve success rate given a larger budget.

Lastly, due to our more strict definition of the attack space, described in (\S\ref{subsec:attack_space}), success rates of \bff{} and \txf{} are lower compared to \newcite{jin2020bert}. To verify the correctness of our attacks, we run \bff{} and \txf{} in their attack space, which uses a larger synonym dictionary, a more permissive function $\Phi$, and does not limit the number of substitutions $D$ and budget $B$. We obtain a similar success rate, close to 100\%.
Nevertheless, we argue our attack space, validated by users to be label-preserving is preferable, and leave standardization of attack spaces through a broad user study to future work.

\subsection{User Study}
\label{subsec:userstudy}

\begin{table}[t]
\centering
\footnotesize
\begin{tabular}{l@{\extracolsep{6pt}~~}c@{~}c@{~~}c@{~~}} 
\Xhline{2\arrayrulewidth}
& Original &  Random &  \bff{} \\ 
\hline
IMDB   & $98.0$   & $98.0$   & $96.0$   \\
BoolQ  & $89.0$   & $91.5$   & $83.5$    \\
SST-2  & $97.0$   & $96.0$   & $94.4$   \\
\Xhline{2\arrayrulewidth}
\end{tabular}
\vskip 0.1in
\caption{Evaluating attack space validity. We show human performance on original examples, random examples, and  examples generated with \bff{}.
}
\label{table:user-study}
\end{table}

Since a model is considered to \emph{not} be robust even if it flips the output label for a single adversarial sample, the validity of adversarial examples in the attack space is crucial. When we examined generated attacks based on prior works, we found many label-flipping attacks. This was especially noticeable when using \bff{} attacks over tasks not evaluated in prior works (see examples in Appendix \ref{app:attack-prior-work}).
In this work, our focus was on evaluating different methods for increasing model robustness, and thus over-constraining the attack space to guarantee its validity was acceptable. We stress that our attack search space is more conservative than prior work, and is a strict subset of prior attack spaces (see Appendix \ref{appendix:attack-space}), leading to higher validity of adversarial examples.

We evaluate the validity of our attack space and the generated adversarial samples with a user study. 
We sample 100/100/50 examples from SST-2/BoolQ/IMDB respectively, and for each example create two adversarial examples: (a) by random sampling (b) using a \bff{} attack. We ask 
25 NLP graduate students to annotate both the original example and the two adversarial ones. Each example is annotated by two annotators and each annotator only sees one version of an example. 
If human performance on random and adversarial examples is similar to the original task, this indicates the attack space is label-preserving.

Table~\ref{table:user-study} shows the results. Human performance on random examples is similar to the original utterances. Human performance on examples generated with \bff{} is only mildly lower than the performance on the original utterances, overall confirming that the attack space is label-preserving.

Ideally, the validity of adversarial exmaples should be as high as the original examples. However, a small degradation in random vs. original is expected since the search space is not perfect, and similarly for BFF since it is targeted at finding adversarial examples. Nevertheless, observed drops were small, showing the advantage in validity compared to prior work. The minor irregularity in BoolQ between random and original is indicative of the noise in the dataset.

\section{Related Work}
\label{sec:related}

Adversarial attacks and robustness have attracted tremendous attention. We discuss work beyond improving robustness through adversarial attacks.

\paragraph{Certified Robustness} is a class of methods that 
provide a mathematical certificate for robustness \cite{dvijotham2018training,gowal2018effectiveness,jia2019certified, huang2019achieving,shi2020robustness}. The model is trained to minimize an upper bound on the loss of the worst-case attack. When this upper bound is low, we get a certificate for robustness against all attacks. While this approach has had success, it struggles when applied to transformers, since upper bounds are propagated through many layers, and become too loose to be practical.

\paragraph{Gradient-based methods}
In a white-box setting, adversarial examples can be generated by performing gradient ascent with respect to the input representation. Gradient-based methods \cite{goodfellow2014explaining,madry2017towards} have been empirically successful \cite{gowal2018effectiveness,ebrahimi2017hotflip}, but suffer from a few shortcomings: (a) they assume access to gradients, (b) they lose their effectiveness when combined with sub-word tokenization, since one cannot substitute words that have a different number of sub-words, and (c) they can generate noisy examples that does not preserve the output label. In parallel to our work, \newcite{guo2021gradient} proposed a gradient-based approach that finds a distribution over the attack space at the token level, resulting in an efficient attack. 

\paragraph{Virtual adversarial training} 
In this approach, one does not generate explicit adversarial examples \cite{zhu2019freelb, jiang2019smart, Li2020TAVATTV, pereira2021targeted}. 
Instead, embeddings in an $\epsilon$-sphere around the input (that do not correspond to words) are sampled, and continuous optimization approaches are used to train for robustness. These works were shown to improve downstream accuracy, but did not result in better robust accuracy. Recently, \newcite{zhou2020defense} proposed a method that does improve robustness, but like other gradient-based methods, it is white-box, does not work well with transformers over sub-words, and leads to noisy samples. A similar approach has been taken by \newcite{si2020better} to generate virtual attacks during training by interpolating offline-generated attacks.

\paragraph{Defense layers}
This approach involves adding normalization layers to the input before propagating it to the model, so that different input variations are mapped to the same representation \cite{wang2019natural,mozes2020frequency,jones2020robust} . While successful, this approach requires manual engineering and a reduction in model expressivity as the input space is significantly reduced. A similar approach \cite{zhou2019learning} has been to identify adversarial inputs and predict the original un-perturbed input.

\paragraph{Pretrained language-models as attacks}
\label{subsec:bert-attack}

In this work, we decouple the definition of the attack-space from the attack strategy itself, which is cast as a search algorithm. This allows us to systematically compare different attack strategies and methods to improve robustness in the same setting. An orthogonal approach to ours was proposed by \newcite{garg2020bae} and \newcite{li2020bert}, who 
used the fact that BERT was trained with the masked language modeling objective to predict possible semantic preserving adversarial perturbations over the input tokens, thereby coupling the definition of the attack space with the attack strategy.
While this approach showed great promise in efficiently generating valid adversarial examples, 
it does not permit any external constraint on the attack space and thus is not comparable to attacks in this work. Future work can test whether robustness transfers across attack spaces and attack strategies by either (a) evaluating the robustness of models trained in this work against the aforementioned works (in their attack space), or (b) combine such attacks with online augmentation to train robust models and compare to the attacks proposed in our work.

\commentout{
\paragraph{Defense layers}
A natural defense against attacks is erecting walls. Multiple works such as \cite{wang2019natural, mozes2020frequency} showed how to add normalization layers on the inputs prior to propagating it through the model. Though somewhat successful, this approach suffers from high level of manual engineering required and a big reduction in the model expressivity as the input space is significantly reduced. A similar approach \cite{zhou2019learning} attempts to identify adversarial inputs and predict the original un-perturbed input that may have been their source. Superficially this approach increase robustness against attacks aimed at the base model, but as the attacker in these works is never presented with the option to probe the complete pipeline, it is not clear it will indeed remain effective if it would. Since it impossible to  expose one model to attackers and another (namely the protected model) to the final query, this is an infeasible direction.  
}

\section{Conclusions}
We examine achieving robustness through discrete adversarial attacks. We find that the popular approach of offline augmentation is sub-optimal in both speed and accuracy compared to random sampling, and that online augmentation leads to impressive gains. 
Furthermore, we propose \bff{}, a new discrete attack based on best-first search, and show that it outperforms past work both in terms of robustness improvement and in terms of attack success rate.

Together, our contributions highlight the key factors for success in achieving robustness through adversarial attacks, and open the door to future work on better and more efficient methods for achieving robustness in natural language understanding.

\section*{Acknowledgements}
We thank Mor Geva, Tomer Wolfson, Jonathan Herzig, Inbar Oren, Yuval Kirstain, Uri Shaham and Omer Levy for their useful comments. This research was partially supported by The Yandex Initiative for Machine Learning, and the European Research Council (ERC) under the European
Union Horizons 2020 research and innovation programme (grant ERC DELPHI 802800).

\bibliography{emnlp2021}
\bibliographystyle{acl_natbib}

\appendix
\clearpage  
\setcounter{page}{0}
\pagenumbering{arabic}
\setcounter{page}{1}

\section{Appendix}
\label{sec:appendix}

\subsection{Experimental Details}
\label{appendix:implmentation-details}

All of the code was written in \emph{python} and is available at \url{https://github.com/Mivg/robust_transformers}. 
The models are trained with the \emph{transformers} library \cite{wolf-etal-2020-transformers}. Whenever \emph{offline augmentation} was used, the resulting adversarial samples were added to the training set and shuffled before training a new model with the same hyper-parameters as the baseline. Thus, the model is trained on $N\times L$ samples where $N$ is the original numbers of samples and $L$ is the number of augmentations added per sample. 
For \emph{online augmentation}, we run two parallel data loaders with different shuffling, each with half the required batch size. We then attack the samples in one batch and concatenate the most successful attack to the other batch. The model is fed with the new constructed batch with identical weighting to the halves. Here, we consider a full epoch when every sample was passed through the model both as a perturbed and as an unperturbed sample. As such, the model is trained on $2N$ samples.
For each dataset, we use the default train-dev split as described in the paper, and report the accuracy on the development set. We train with hyper-parameters as described below:
\begin{enumerate}[leftmargin=0pt,itemsep=0pt,topsep=0pt]
    \item[] \textbf{SST-2}: We fine-tuned a pre-trained cased \bertb \cite{devlin2018bert} with \emph{max seq length}$=128$ over Nvidia Titan XP GPU for three epochs with batch size of 32 and learning rate of $2e-5$.
    \item[] \textbf{IMDB}: We fine-tuned a pre-trained cased \bertb \cite{devlin2018bert} with \emph{max seq length}$=480$ over Nvidia Titan XP GPU for three epochs with batch size of 48 and learning rate of $2e-5$.
    \item[] \textbf{BoolQ}: We fine-tuned a pre-trained \robertal \cite{liu2019roberta} for BoolQ with \emph{max seq length}$=480$ over Nvidia GTX 3090 GPU for three epochs with batch size of 48 and learning rate of $1e-5$.
\end{enumerate}

For each parameter choice reported in Table~\ref{table:results}, we ran three different experiments with different random initialization, and reported the mean results. The respective standard deviations are given in Table~\ref{table:results-std}.
To finetune the models using the FreeLB \cite{zhu2019freelb} method, we adapted the implementation from \url{https://github.com/zhuchen03/FreeLB} and used the following parameters: 
\begin{enumerate}[leftmargin=0pt,itemsep=0pt,topsep=0pt]
    \item[] \textbf{SST-2}: $\text{init-magnitude}=0.6$, $\text{adversarial-steps}=2$, $\text{adversarial-learning-rate}=0.1$ and $l_2$ norm with no limit on the norm.
    \item[] \textbf{IMDB}:  $\text{init-magnitude}=0.2$, $\text{adversarial-steps}=4$, $\text{adversarial-learning-rate}=0.2$ and $l_2$ norm with no limit on the norm.
    \item[] \textbf{BoolQ}:  $\text{init-magnitude}=0.2$, $\text{adversarial-steps}=4$, $\text{adversarial-learning-rate}=0.2$ and $l_2$ norm with no limit on the norm.
\end{enumerate}

\begin{table}[t]
\centering
\footnotesize

\resizebox{\columnwidth}{!}{
\begin{tabular}{@{\extracolsep{1.5pt}}lcccccc@{}}

\Xhline{2\arrayrulewidth}

\multirow{2}{*}{\textbf{Model}} & \multicolumn{3}{c}{\textbf{Accuracy}} & \multicolumn{3}{c}{\textbf{Robust Accuracy}}  \\\cline{2-4}\cline{5-7}

& {\footnotesize \textbf{SST-2}} & {\footnotesize \textbf{IMDB}} & {\footnotesize \textbf{BoolQ}} & {\footnotesize \textbf{SST-2}} & {\footnotesize \textbf{IMDB}} & {\footnotesize \textbf{BoolQ}} \\ \cline{2-2}\cline{3-3}\cline{4-4}\cline{5-5}\cline{6-6}\cline{7-7}

Baseline          & $\pm0.1$ & $\pm0.1$ & $\pm1.3$ & $\pm0.4$ & $\pm0.6$ & $\pm0.9$ \\
\freelb{}        & $\pm0.2$ & $\pm0.1$ & $\pm0.4$ & $\pm0.5$ & $\pm1.0$ & $\pm1.1$ \\
\hline\hline
\randoff-1        & $\pm0.3$ & $\pm0.1$ & $\pm1.8$ & $\pm0.5$ & $\pm1.4$ & $\pm1.8$ \\
\randoff-4        & $\pm0.7$ & $\pm0.1$ & $\pm0.5$ & $\pm0.6$ & $\pm1.9$ & $\pm0.5$ \\
\randoff-8        & $\pm0.2$ & $\pm0.1$ & $\pm0.8$ & $\pm0.7$ & $\pm2.1$ & $\pm0.8$ \\
\randoff-12       & $\pm0.6$ & $\pm0.1$ & $\pm1.0$ & $\pm0.5$ & $\pm1.4$ & $\pm1.0$ \\
\hline
\txfoff{}         & $\pm0.6$ & $\NA$ & $\NA$ & $\pm0.3$ & $\NA$ & $\NA$ \\

\bffoff{}         & $\pm0.3$ & $\NA$ & $\pm0.3$ & $\pm0.3$ & $\NA$ & $\pm1.8$ \\
\hline\hline
\randon{}         & $\pm0.1$ & $\NA$ & $\NA$ & $\pm0.3$ & $\NA$ & $\NA$ \\
\txfon{}          & $\pm0.0$ & $\NA$ & $\NA$ & $\pm0.3$ & $\NA$ & $\NA$ \\
\bffon{}          & $\pm0.5$ & $\NA$ & $\NA$ & $\pm0.6$ & $\NA$ & $\NA$ \\
\Xhline{2\arrayrulewidth}
\end{tabular}
}
\caption{Standard deviation on the experiments reported in Table~\ref{table:results}. Missing cells indicate a single-run was used due to the long training time.}
\label{table:results-std}
\end{table}

\begin{figure}[t]
\begin{center}
\centerline{\includegraphics[trim={0 0 0 0.55cm},clip,width=\linewidth]{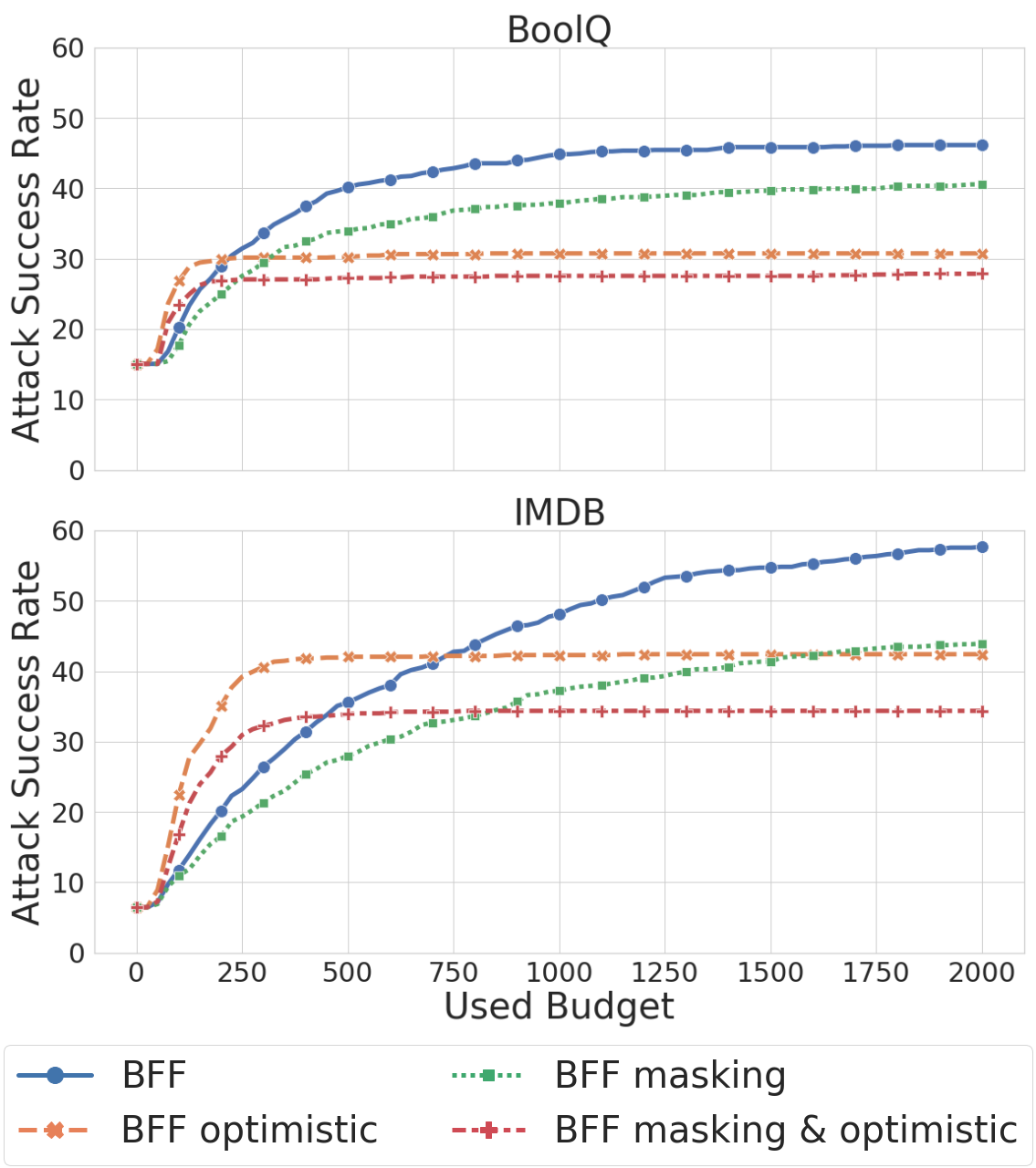}}
\caption{Success rate of different attacks against BoolQ/IMDB \baseline{} as a function of the budget.}
\label{fig:appendix-ablations}
\end{center}
\end{figure}

\paragraph{BFF implementation} 
For the factorization phase of \bff{}, we use $\tau \sim \syn{w}$ with uniform sampling. We find that while using an out-of-vocabulary masking token is useful to compute a word salience, it is less suitable here as we are interested in the model's over-sensitivity to perturbations in the exact phrasing of the word. Also, in contrast to \txf{} which is optimistic and factorizes the attack space only once, \bff{} factorizes the space after every step.
Namely, \emph{Optimistic greedy search} plans the entire search path by evaluating all permissible single-word substitutions. Let $x_{w_i \rightarrow w}$ denote the utterance $x$ where the word $w_i$ is replaced with a synonym $w \in \syn{w_i}$. The optimistic greedy algorithm
scores each word $w_i$ in the utterance with $s(w_i) \coloneqq \min_{w \in \mathit{syn}(w_i)} s_A(x_{w_i \rightarrow w})$, that is, the score of a word is the score for its best substitution, and also stores this substitution. Then, it sorts utterance positions based on $s(w_i)$ in ascending order, which defines the entire search path: In each step, the algorithm moves to the next position based on the sorted list and uses the best substitution stored for that position.
Fig.~\ref{fig:appendix-ablations} shows the benefit from each of those modifications.

\begin{figure}[t]
\begin{center}
\centerline{\includegraphics[width=\linewidth]{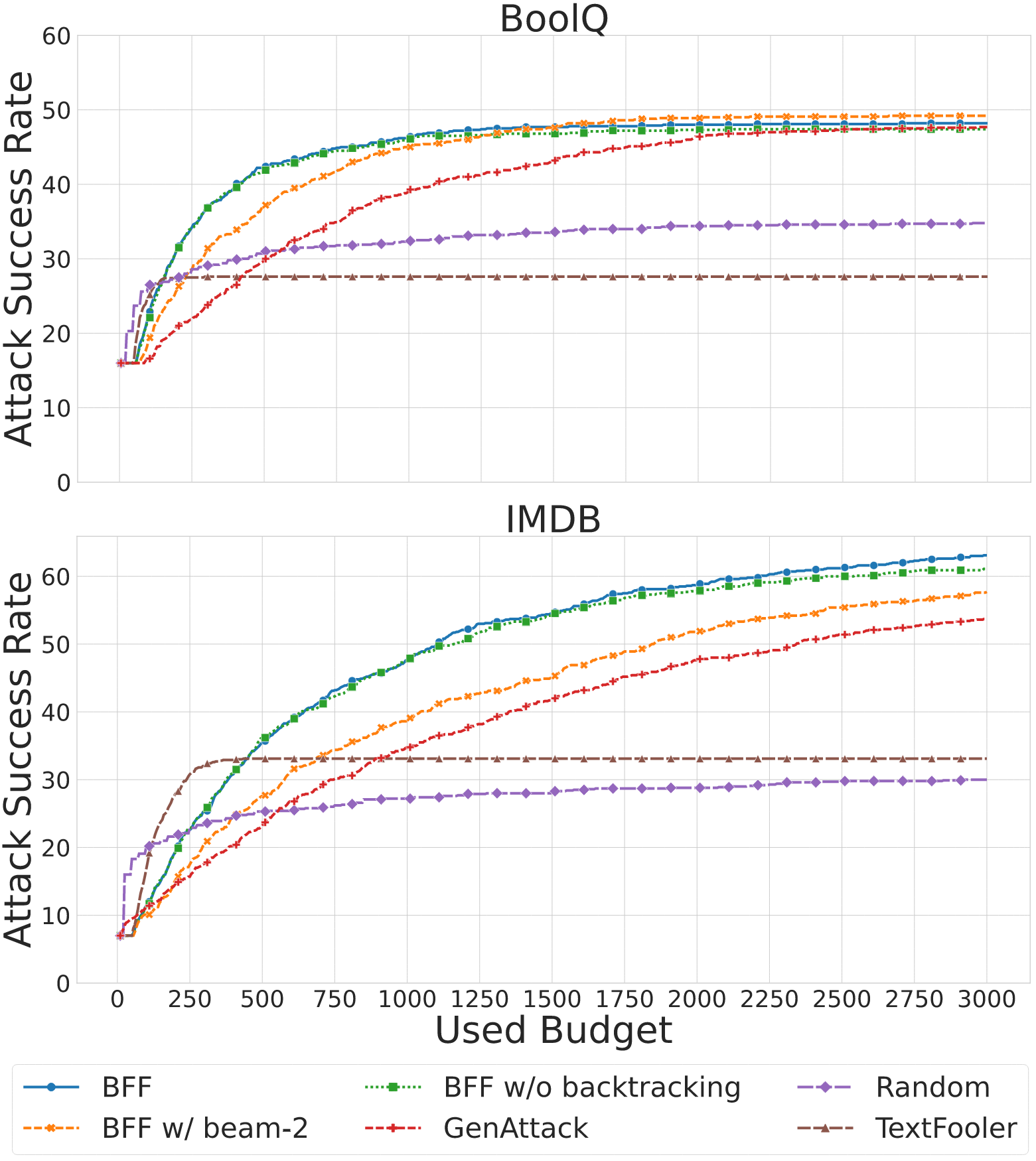}}
\caption{Success rate  of different attacks against BoolQ/IMDB \baseline{} as a function of the budget.}
\label{fig:attack-results-budget3000}
\end{center}
\end{figure}

\paragraph{Budget Effect}
Intuitively, higher budgets better approximate an exhaustive search and thus the robustness evaluation as an upper bound should approach its true value. However, due to lack of backtracking in some of the attack strategies, they may plateau early on. In this work, we used $B=1000$ for all training phases and $B=2000$ for the robustness evaluation. Empirically, this gives a good estimate on the upper bound of model's robust accuracy, while constraining the computational power needed for the experiments. A natural question is how much tighter the bounds may be if a larger budget is given. Fig.~\ref{fig:attack-results-budget3000} depicts an evaluation of strategies' success-rates over the same models as in Fig.~\ref{fig:attack-results} with a larger budget. As can be seen, while the \textsc{Random} attack and \txf{} plateau, \bff{} variants as well as \textsc{GenAttack} are able to exploit the larger budget to fool the model in more cases. This is especially true in IMDB where the search space is considerably larger. We expect this trend of tighter bounds to continue with ever larger budgets, though we note that the rate of improvements decreases with budget and that the ranking between strategies remains unchanged. Therefore, we conclude that drawing conclusions about strategies comparison and robustness improvements by evaluating with a budget of $2,000$ suffices.

\subsection{Attack Space Implementation Details}
\label{appendix:attack-space}

As described in \S\ref{subsec:attack_space}, we use the synonyms dictionary defined by \newcite{alzantot2018generating}. In particular, we use the pre-computed set of those synonyms given by \newcite{jia2019certified} as our bases for \syn{w}. We pre-process the entire development and training data and store for each utterance, the set $\mathit{Syn}_{\Phi}(w)$ and avoid the need to employ large language models during training and robustness evaluation.
For every word in an utterance $w_i \in x$, and for every $\bar{w}_i \in \syn{w_i}$ we evaluate $\Phi(w_i,\bar{w}_i)$ as follows:
\begin{enumerate}[leftmargin=5.0pt,itemsep=0pt,topsep=0pt]
    \item With the same sequences as above, we validate that $POS(w_i)\equiv POS(\bar{w}_i)$ according to spaCy's \cite{spacy} POS tagger.
    \item With a window of size 101, we validate that $\text{PPL}(x)/\text{PPL}(\bar{x})\ge0.8$ where $\text{PPL}(\cdot)$ is the perplexity of the sequence as given by a pre-trained GPT-2 model \cite{radford2019language}
    \item For BoolQ only, we also use spaCy's POS tagger to tag all content words (namely \emph{NOUN}, \emph{PROPN}, \emph{ADV}, and \emph{ADJ}) in the question. We then restrict all those words from being perturbed in the passage.
    \item Following \newcite{jin2020bert}, we take a window of size 15 around the word, and validate with \textsc{USE} \cite{cer2018universal} that the semantic similarity between the unperturbed sequence $(w_{i-7},\dots,w_i,\dots,w_{i+7})$ and the perturbed sequence $(w_{i-7},\dots,\bar{w}_i,\dots,w_{i+7})$ is at least 0.7.
\end{enumerate}

\subsection{Genetic Attack Implementation Details}
\label{appendix:gen-attack-implmentation-details}

Our implementation of \emph{Gen-Attack} presented by \newcite{alzantot2018generating} was based on \url{https://github.com/nesl/nlp_adversarial_examples/blob/master/attacks.py} and used our attack space rather than the original attack space presented there. For evaluation, we used the distribution hyperparameters as defined by the paper. Namely, \textit{population-size}: $p := 20$, \textit{maximum generations}: $g := 100$ and $\text{softmax-temperature}=0.3$. Note we did not need to limit the number of candidate synonyms considered as this was already done in the attack space construction.  However, we have made two modifications to the original algorithm in order to adapt to our settings.

\paragraph{Maximal modification constraints} While the original algorithm presented by \newcite{alzantot2019genattack} contained a \textit{clipping} phase where mutated samples where clipped to match a maximal norm constraint, the adapted version for discrete attacks presented in \newcite{alzantot2018generating} did not. As we wish to limit the allowed number of perturbation for any single input utterance and the crossover phase followed by the \textit{perturb} sub-routine can easily overstep this limit, a post-perturb phase was added. Namely, in every generation creation, after the crossover and mutation (i.e. perturb) sub-routines create a candidate child, if the total number of perturbed samples exceeds the limit, we randomly uniformly revert the perturbation in words until the limit is reached. This step introduced another level of randomness into the process. We experimented with reverting based on the probability to be replaced as used in the \textit{perturb} sub-routine, but this resulted in sub-par results.

\paragraph{Improved Efficiency} In addition to estimating the fitness function of each child in a generation which requires a forward pass through the attacked model, \newcite{alzantot2018generating} also used a greedy step in the \textit{perturb} sub-routine to estimate the fitness of each synonym mutation for a chosen position. This results in an extremely high number of forward passes through the model, specifically $\mathcal{O}(g\cdot p\cdot (k+1))$ which is orders of magnitude larger than our allowed budget of 2000. However, many of the passes are redundant, so by utilizing caching to previous results, the attack strategy can better utilize its allocated budget, resulting in significantly better success rate in with better efficiency.
\subsection{Attack Space in Prior Work}
\label{app:attack-prior-work}

Examining the attack space proposed in \newcite{jin2020bert}, which includes a larger synonym dictionary and a different filtering function $\Phi(\cdot)$,
we observe that many adversarial examples are difficult to understand or are not label-preserving. Table~\ref{table:bad-utterances} shows examples from an implementation of the attack space of the recent \textsc{TextFooler} \cite{jin2020bert}. We observe that while in IMDB the labels remain mostly unchanged, many passages are difficult to understand. Moreover, we observe frequent label flips in datasets such as in SST-2 example, as well as perturbations in BoolQ that leave the question unanswerable.

\begin{table*}[t]
\centering
\small
\begin{tabular}{p{0.95\linewidth}}
\Xhline{2\arrayrulewidth}
\textbf{Passage}: Table of prime factors -- The number 1 is called a unit. It has no \pert{incipient} [\emph{prime}] factors and is neither \pert{fiirst} [\emph{prime}] nor composite. \\
\textbf{Question}: is 1 a prime factor of every number \\
\textbf{Answer}: False \\
\cdashline{1-1}
\textbf{Passage}: Panama Canal -- The \pert{nouvelle} [\emph{new}] locks \pert{commences} [\emph{opened}] for commercial \pert{vehicular} [\emph{traffic}] on 26 June 2016, and the first \pert{yacht} [\emph{ship}] to \pert{intersecting} [\emph{cross}] the canal using the third set of locks was a modern New Panamax vessel, the Chinese-owned container \pert{warships} [\emph{ship}] Cosco Shipping Panama. The original locks, now over 100 \pert{centuries} [\emph{years}] old, \pert{give} [\emph{allow}] \pert{engineer} [\emph{engineers}] \pert{best} [\emph{greater}] access for maintenance, and are \pert{hoped} [\emph{projected}] to continue \pert{workplace} [\emph{operating}] indefinitely.  \\
\textbf{Question}: is the old panama canal still in use\\
\textbf{Answer}:  True \\
\cdashline{1-1}
\textbf{Passage}:  Chevrolet Avalanche -- The Chevrolet Avalanche is a four-door, five or \pert{eight} [\emph{six}] \pert{commuter} [\emph{passenger}] \pert{harvest} [\emph{pickup}] \pert{trucking} [\emph{truck}] \pert{stocks} [\emph{sharing}] GM's long-wheelbase \pert{frame} [\emph{chassis}] used on the Chevrolet Suburban and Cadillac Escalade ESV. Breaking with a long-standing tradition, the Avalanche was not \pert{affordable} [\emph{available}] as a GMC, but only as a Chevrolet. \\
\textbf{Question}:   is there a gmc version of the avalanche \\
\textbf{Answer}:  False \\
\hline\hline
\textbf{Sentence}: I've been waiting for this movie for SO many years! The best part is that it \pert{decedent} [\emph{lives}] up to my visions! This is a MUST SEE for any Tenacious D or true Jack Black fan. It's just \pert{once} [\emph{so}] great to see JB, KG and Lee on the big screen! It's not a \pert{authentic} [\emph{true}] story, but who cares. The D is the greatest band on earth! I had the soundtrack to the movie last week and \pert{heeded} [\emph{listened}] to it non-stop. To see the movie was \pert{unadulterated} [\emph{pure}] bliss for me and my hubby. We've both met Jack and Kyle after 2 different Tenacious D concerts and also saw them when they toured with Weezer. We left that concert after the D was done playing. Nobody can top their show! Long live the D!!! :D\\
\textbf{Answer}: True \\
\cdashline{1-1}
\textbf{Sentence}: Sweet, \pert{kidding} [\emph{entertaining}] tale of a young 17 1/2 year old boy, controlled by by an overbearing religious mother and withdrawn father, and how he finds himself through his work with a retired, eccentric and tragic actress. Very \pert{better} [\emph{well}] acted, especially by Julie Walters. Rupert Grint plays the role of the teenage boy well, showing his talent will last longer than the Harry Potter series of films. Laura Linney plays his ruthlessly strict mother without a hint of redemption, so there's no room to like her at all. But the film is a \pert{awfully} [\emph{very}] \pert{antics} [\emph{entertaining}] film, made well by the British in the style of the likes of Keeping Mum and Calendar Girls. \\
\textbf{Answer}: True \\
\cdashline{1-1}
\textbf{Sentence}: Enormous \pert{adjourned} [\emph{suspension}] of disbelief is required where Will's "genius" is concerned. Not just in math--he is also very well \pert{reads} [\emph{read}] in economic history, able to out-shrink several shrinks, etc etc. No, no, no. I don't buy it. While they're at it, they might as well have him wearing a big "S" on his chest, flying faster than a jet plane and stopping bullets.$<$br $/>$ $<$br $/>$Among other problems...real genius (shelving for the moment the problem of what it really is, and whether it deserves such mindless homage) doesn't simply appear /ex nihilo/. It isn't ever so multi-faceted. And it is very \pert{virtually} [\emph{rarely}] \pert{appreciates} [\emph{appreciated}] by contemporaries.<br /><br />Better to have made Will a basketball prodigy. Except that Damon's too short. \\
\textbf{Answer}: False \\
\hline\hline
\textbf{Sentence}: it proves quite \pert{unconvincing} [\emph{compelling}] as an intense , brooding character study . \\
\textbf{Answer}: True\\
\cdashline{1-1}
\textbf{Sentence}: an \pert{sensible} [\emph{unwise}] amalgam of broadcast news and vibes . 	an sensible amalgam of broadcast news and vibes . \\
\textbf{Answer}: False \\ 
\cdashline{1-1}
\textbf{Sentence}: if you dig on david mamet 's mind tricks ... rent this movie and \pert{iike} [\emph{enjoy}] ! \\
\textbf{Answer}: True \\
\Xhline{2\arrayrulewidth}
\end{tabular}
\caption{Examples of adversarial examples, which are difficult to understand or not label-preserving, found for BoolQ/IMDB/SST-2 with the attack space from \cite{jin2020bert}. In \pert{bold} are the substituting words and in \emph{brackets} the original word.
}
\label{table:bad-utterances}
\end{table*}

\end{document}